MaskIt: Masking for efficient utilization of incomplete public datasets for training deep learning models

Ankit Kariryaa[1]

1 HCI Group, University of Bremen, 28359 Bremen, Germany. E-Mail: kariryaa@uni-bremen.de.

**ABSTRACT**

A major challenge in training deep learning models is the lack of high quality and complete datasets. In the paper, we present a masking approach for training deep learning models from a publicly available but incomplete dataset. For example, city of Hamburg, Germany maintains a list of trees along the roads, but this dataset does not contain any information about trees in private homes and parks. To train a deep learning model on such a dataset, we mask the street trees and aerial images with the road network. Road network used for creating the mask is downloaded from OpenStreetMap, and it marks the area where the training data is available. The mask is passed to the model as one of the inputs and it also coats the output. Our model learns to successfully predict trees only in the masked region with 78.4% accuracy.

**MAIN**

Biodiversity and the wild population of plants and animals are rapidly decreasing throughout the world [4]. Recent articles have suggested that the sixth mass extension on earth is underway [2]. This phenomenon is not just limited to the large animals, similar reports have emerged for insects [6] and plant species. Human impact is primarily blamed for this crisis [9]. For example, industrial-scale use of pesticides and insecticides in farming is often described as disastrous for the insects. Recent reports have suggested that even in protected areas the insect population is dwindling to as low as 25% in the last 25 years [6]. Since insects are the prey for birds and wild animals, the decrease in their population affects various other species as well. Large-scale deforestation, loss of habitat for wild animals and intrusion into nature is also seen as main cause of the current COVID-19 pandemic.

As a primary step in monitoring biodiversity at a global scale, a system is required to determine the type of tree and plant species in various environments. In the past, satellite imagery has been successfully applied to monitor the forests dynamics [7]. Recent advancements in deep learning along with increased availability of high-resolution satellite imagery (<1 m/pixel), have opened up the door of possibilities for detecting individual tree, plant or crop species with relatively high accuracy [11]. However, a common problem for such a system is the lack of high-quality training data. Public agencies such as city official and forest service often maintain valuable records of public attributes such as road signs, parking areas and trees. However, these datasets are not designed for the use case of the training deep learning models in mind. These datasets are often limited to public areas such as roads and public parks, thus training deep learning models on such datasets can be a challenging task. We here propose a masking approach with which one can efficiently train a deep learning model from incomplete datasets.

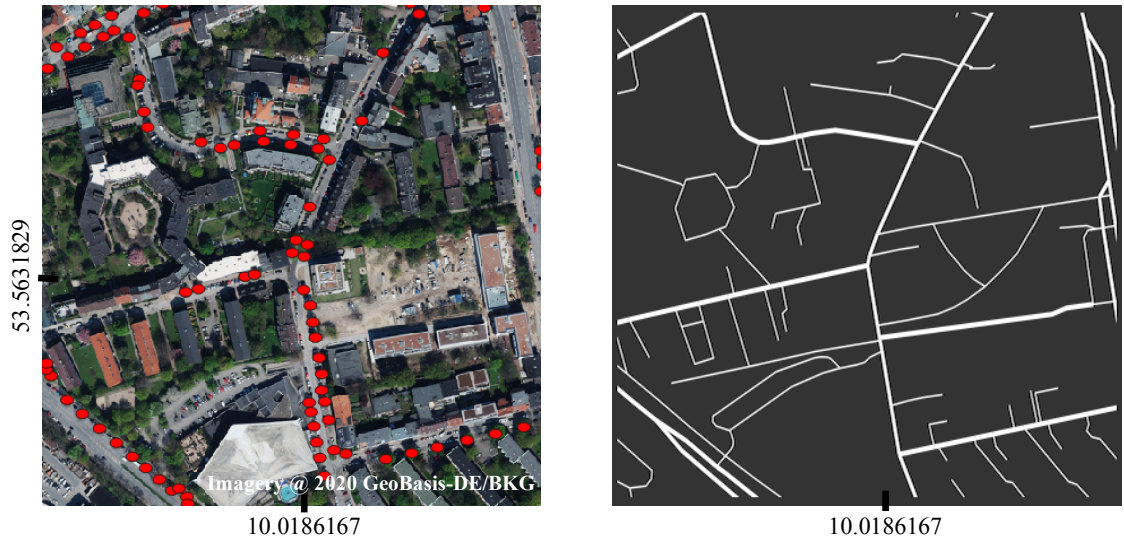

Figure 1. Left panel: Aerial image of an area in Hamburg with street trees marked in red. Right panel: Road network of the same area, while this image shows all the roads, in our case we only used the main roads for creating the mask.

**Dataset**

We use the aerial images and street tree dataset from Hamburg, Germany for training our model. The aerial images had 3 channels (RGB) with 0.2m/pixel resolution and they were downloaded from the Geoportal.de[1]. As seen in the left panel of Figure 1, the individual features such as trees and cars parked on the streets are visible to the human eyes.

Authority for Environment and Energy of the city of Hamburg maintains a list of all street trees[2]. The dataset contains various attributes of individual trees such as location, height, width, species, age, and condition. However, as the name suggests, this information is only limited to the trees on the street of Hamburg and do not contain any information about the trees in private areas and parks. Left panel of Figure 1 shows a sample of the information available in this dataset. The trees are marked in red ovals.

To make use of this dataset, we create a mask based upon the street network of Hamburg. We use the OSMnx python package [3] to download the street networks from OpenStreetMap. The downloaded road network is then transformed and drawn on the aerial image with the help of Rasterio library [5]. We only use the main drivable roads for creating the road mask and add a buffer of 5m on both sides to cover the areas next to roads. We manually annotated 1371 trees crowns along the road network. The tree crowns can also be created from the street tree data, albeit with some noise. The goal is to predict these annotations using the aerial images and road network mask as the input.

---

[1] https://www.geoportal.de/portal/main/
[2] http://suche.transparenz.hamburg.de/dataset/strassenbaumkataster-hamburg7

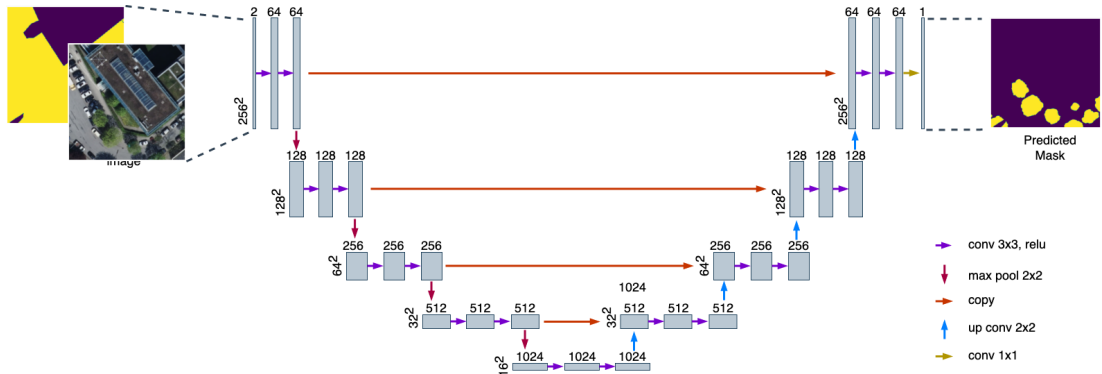

Figure 2: The U-Net architecture. As input we pass the road mask and the aerial image. The model is trained to the predict the tree only in the masked region.

**Mapping trees**

We use a state-of-the-art deep learning model to detect tree crowns in the input images. Our deep learning model was based on the U-Net architecture [10] which was developed for medical image segmentation and is one of the most widely used architectures for semantic segmentation tasks. It is argued that due to the lateral connections from the start to the end layers in the network, the network well preserves the syntactic information in the image. It is known to generalize well on relatively small datasets [8]. Figure 2 shows the architecture of the U-Net model. As input to the model, we pass patches of aerial image and the street network mask, thus a total of 4 channels (RGB + mask).

**Training**

For training the models we extracted 2888 patches of 256*256*4-pixel from a 1km* 2km annotated area. The patches were sequentially extracted with a step of 128 pixels in both directions. They were then randomly divided into 60% training, 20% validation and 20% test patches. The patches were zero padded, if they did not fall completely in the annotated area.

**Results**

Figure 3 shows the result of our approach. The model achieved a per pixel accuracy of 78.4% and successfully detects trees in the aerial imagery. The mask can be seen as the AND operator and the model learns to predict only those pixels where corresponding value in mask is equal to one. The models can be used to predict trees in areas beyond the road network by simply passing a mask of ones. The model accuracy can be further improved using higher quality dataset and data augmentation. The clumped trees can be separated using a weighted loss on the edges [10].

**Discussion**

Here we show that the masking technique can be used to effectively train on incomplete datasets. However, this approach also has some shortcomings. For example, in our case the model is only trained with trees along the streets, and the model did not observe trees in the other conditions such as in a grass field, along a water body or in a park. Thus, its accuracy may be lower in these conditions where background is different than a paved road or a building. Indeed, we observe that model has trouble distinguishing grass patches from trees (see last row in the Figure 3). A solution to this problem could be to use an additional training dataset covering these conditions. The additional dataset can be oversampled or given extra weight for effective training. While this paper shows a relatively simple use of this approach, which can also be achieved through other methods such as, through a logical conjunction (binary and operator) between the input image and the mask in the preprocessing step or by replacing all the values in the masked area in the input image with a fixed value, which would eliminate the need of passing the mask as an input to the deep learning model. However, the masking approach offers the advantage that it has applicability in much more complex scenarios, for example, using this approach we can train a model to predict tree characteristics (e.g. age, species and height) in the masked areas while segmenting the trees in whole image.

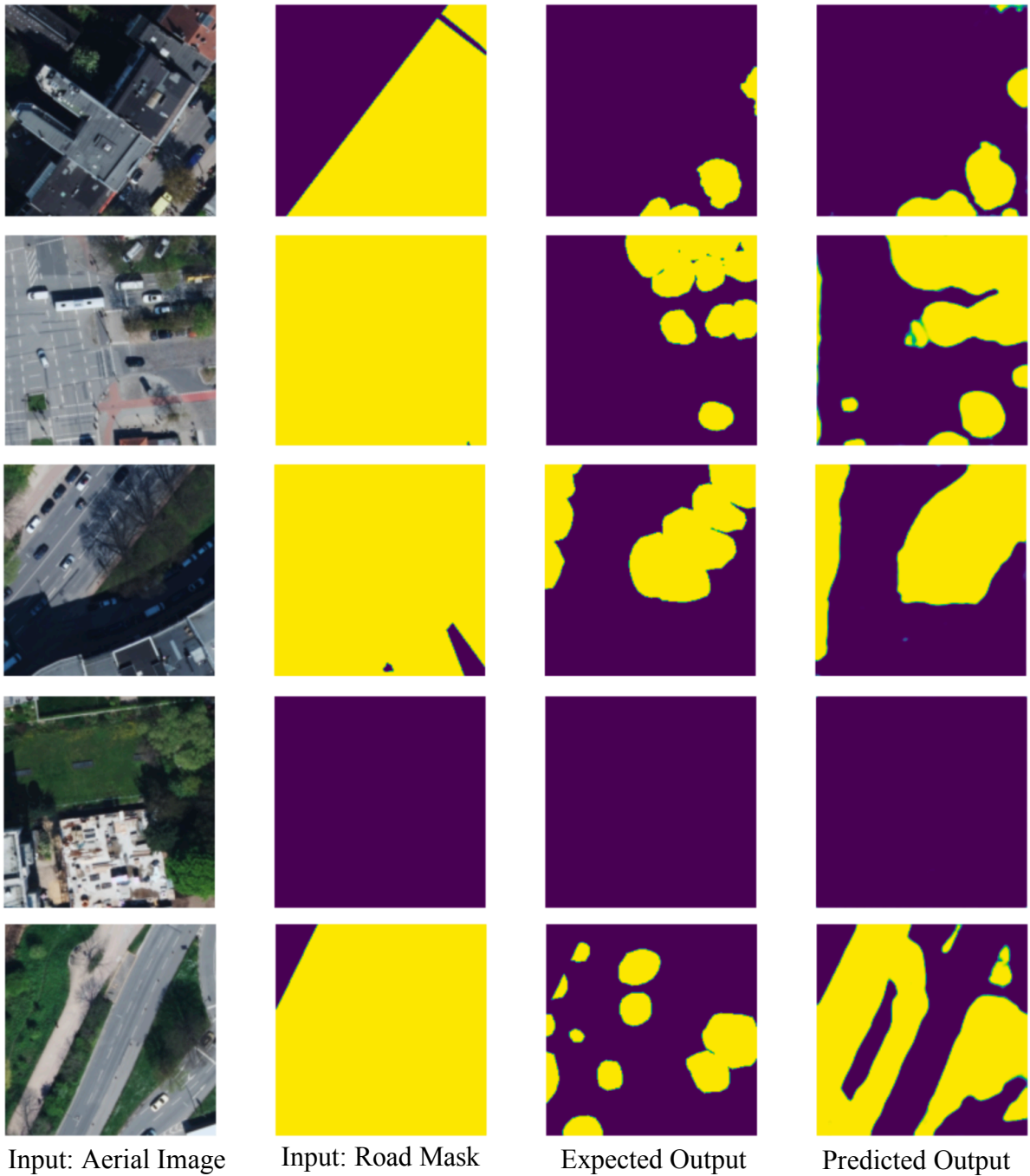

Figure 3: Predicting trees. The first two columns show the model input, the third column shows the expected output and the final column shows the predicted output. The model predicted trees with 78.4% accuracy. The yellow marks the masked area along the road network (second column). The tree class in the expected and predicted output is also shown in yellow. If the mask is empty, the model does not predict in that area as seen in the fourth row. Since the model did not observe trees with grass in the background, its performance is lower in such conditions (for example the last row).

We believe, that in the future this approach can contribute to the overall effort of mapping individual trees. In the future, the greatest opportunity that deep learning and satellite imagery might offer is for large scale citizen science platforms. One can imagine a global platform for monitoring biodiversity, where the data is populated by deep learning pipelines and further refined by citizen scientists. In the future, these platforms may play a crucial role in tackling global challenges including climate change, extinction of species, and continuously shrinking biodiversity. This notion is also shared by several researchers in the field of remote sensing and sustainability, where the number of calls for research in this direction has been growing over the year [1].

**Code availability**
The tree detection framework based on U-Net is made publicly available at https://gitlab.com/Kariryaa/maskit. Please contact the author for support and more information.

**Acknowledgements**
I would like to thanks Sanjeev Sharma for discussions on this idea, Abida Sultana for help with the annotations, Gian-Luca Savino for help with OSMnx library, Daniel Diethei for reviewing the manuscript, Tetiana Gren for support during the project and Johannes Schöning for the general guidance. This research was supported in part by the Volkswagen Foundation through a Lichtenberg Professorship.